\documentclass[10pt,journal,final]{ieeetran}

\IEEEoverridecommandlockouts

\usepackage{multirow} 
\usepackage{booktabs}
\usepackage{makecell}

\usepackage{cite}
\usepackage{graphicx}

\usepackage{amsmath}
\usepackage{listings}
\usepackage{amsthm}
\usepackage{amsmath}
\usepackage{amssymb}
\usepackage{subfigure}
\usepackage{graphicx}

\usepackage[table]{xcolor}
\usepackage{fancyhdr}
\usepackage{footnote}
\usepackage{lastpage}
\usepackage{indentfirst}
\usepackage{multirow}
\usepackage{authblk}
\usepackage{comment}
\usepackage{url}
\usepackage{float}
\usepackage{cite}
\usepackage{caption}
\usepackage[colorlinks,linkcolor=red,anchorcolor=blue,citecolor=green,backref=page]{hyperref}

\begin{document}
\title{UIEC\^{}2-Net: CNN-based Underwater Image Enhancement Using Two Color Space}

\author[1]{Yudong~Wang}
\author[1]{Jichang~Guo\thanks{The work described in this paper was fully supported by a grant from the National Natural Science Foundation of China (No.61771334) (\emph{Corresponding author: Jichang~Guo})}}
\author[1]{Huan~Gao}
\author[1]{Huihui~Yue}
\affil[1]{School of Electrical and Information Engineering, Tianjin University, Tianjin, 300072, China}

\maketitle
\begin{abstract}
Underwater image enhancement has attracted much attention due to the rise of marine resource development in recent years. Benefit from the powerful representation capabilities of Convolution Neural Networks(CNNs), multiple underwater image enhancement algorithms based on CNNs have been proposed in the last few years. However, almost all of these algorithms employ RGB color space setting, which is insensitive to image properties such as luminance and saturation. To address this problem, we proposed Underwater Image Enhancement Convolution Neural Network using 2 Color Space (UICE\^{}2-Net) that efficiently and effectively integrate both RGB Color Space and HSV Color Space in one single CNN. To our best knowledge, this method is the first to use HSV color space for underwater image enhancement based on deep learning. UIEC\^{}2-Net is an end-to-end trainable network, consisting of three blocks as follow: a RGB pixel-level block implements fundamental operations such as denoising and removing color cast, a HSV global-adjust block for globally adjusting underwater image luminance, color and saturation by adopting a novel neural curve layer, and an attention map block for combining the advantages of RGB and HSV block output images by distributing weight to each pixel. Experimental results on synthetic and real-world underwater images show the good performance of our proposed method in both subjective comparisons and objective metrics. The code are available at \url{https://github.com/BIGWangYuDong/UWEnhancement}
\end{abstract}

\section{Introduction}
Nowadays, exploration and utilization of marine resources has become the strategy center in international community. It is very important to make use of these images because underwater images carry plenty of information. However, due to the impact of backscatter in far distances, light selective absorption and scattering in water, the raw underwater images usually suffer from low-quality, including low contrast and brightness, color deviations, blurry details and uneven bright speck. The purpose of underwater image enhancement is to obtain higher quality and clearer underwater images, so as to make more effective use of image information. It can make full use of images information and has been widely used to promote numerous engineering and high-level research tasks such as underwater fish detection, shipwreck detection, underwater archaeology, \textit{etc}.
\captionsetup[figure]{name={Figure}}
\begin{figure}[htbp]
  \centering
  \setlength{\belowcaptionskip}{-10pt}
  \setlength{\abovecaptionskip}{0pt}
  \subfigure{
  \includegraphics[width=0.3\linewidth]{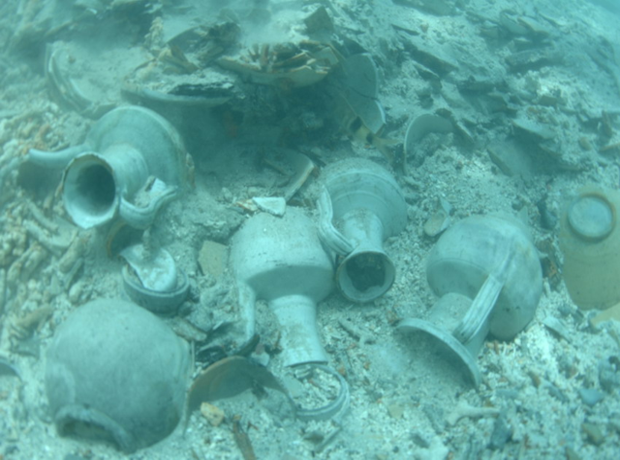}
  \includegraphics[width=0.3\linewidth]{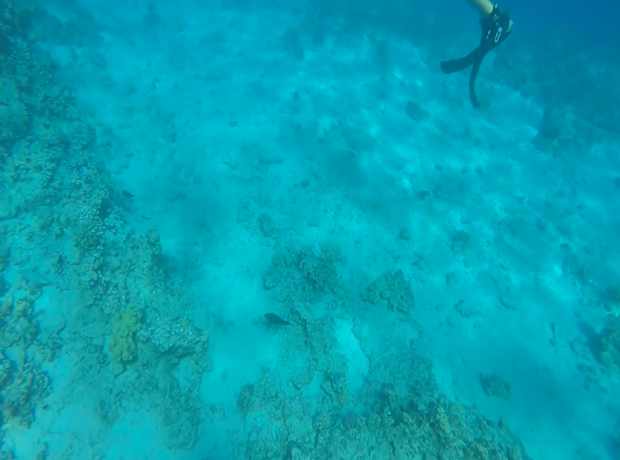}
  \includegraphics[width=0.3\linewidth]{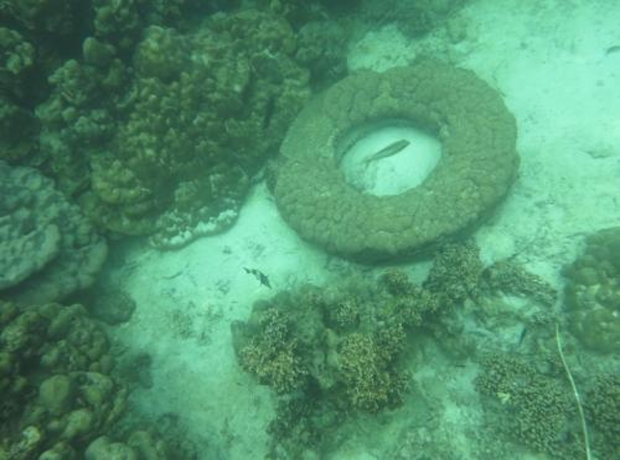}
  }

  \subfigure{
  \includegraphics[width=0.3\linewidth]{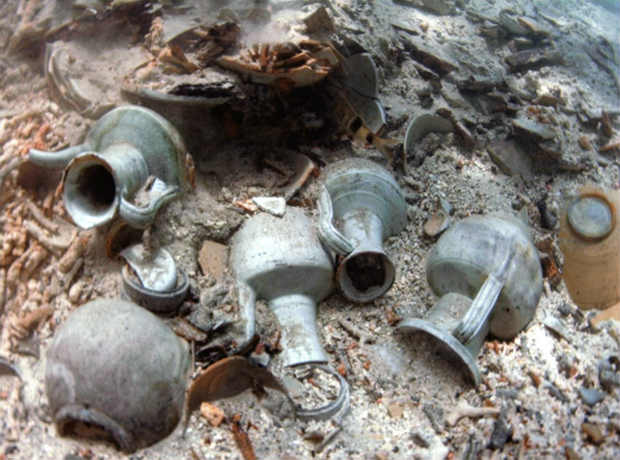}
  \includegraphics[width=0.3\linewidth]{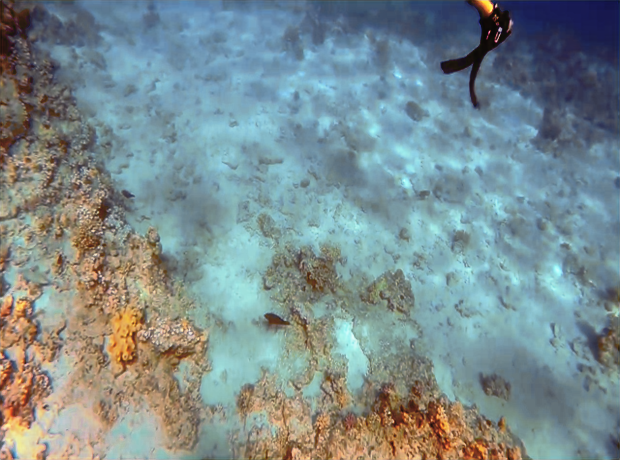}
  \includegraphics[width=0.3\linewidth]{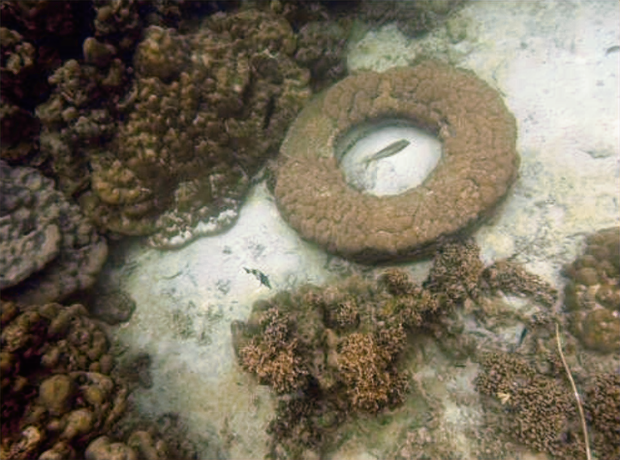}
  }
  \caption{Samples of real-world underwater image enhancement by UIEC\^{}2-Net. Top row: raw underwater images from DUIE datasets \cite{li2019underwater}; Bottom row: the corresponding results of our model.}
  \label{fig1}
\end{figure}

Early underwater image enhancement methods often directly use traditional techniques(\textit{e.g.} histogram equalization based methods,and physical-based methods\cite{lcy2016tip,lcy2017prl}). However, these algorithms have certain limitations: the classic in-air image enhancement methods\cite{hitam2013mixture,rahman1996woodell} are often not outstanding enough in processing underwater image. Most physical model-based methods\cite{li2016singles,li2016single} are based on a simplified underwater image formation model\cite{chiang2011underwater}, which is inaccurate because it assume many parameters. This leads to get not good enough results for some types of underwater images and is only suitable for underwater scenes under certain circumstances. In the past few decades, deep learning technology has developed rapidly, which has brought a series of breakthroughs to various of computer vision and image processing tasks\cite{lecun2015deep,lcy2019tgrs,lcy2019tip1,lcy2018spl,li2020nui,li2020rgb,lcy2020tmm,guo2020zero}. Affected by deep learning, many researchers began to use CNN-based and GAN-based algorithms to enhance underwater image. The goal of CNN-based algorithms is to be faithful to the original underwater image, while the core of GAN-based algorithms is to improve the perceptual quality of the entire underwater image. However, the current underwater image enhancement still fails to observe attractive performance. This might be attributable to the fact that most deep learning-based underwater image enhancement algorithms almost all used RGB Color Space. The RGB color space can deal with the scattering problem and improve the underwater image issues such as color deviations, some issues of quality degradation such as low contrast, saturation and brightness cannot get further improvement due to the RGB color space cannot directly reflect some important parameters of the image.

\begin{figure*}[htbp]
  \centering
  \setcaptionwidth{6.8in}
  \includegraphics[width=0.9\linewidth]{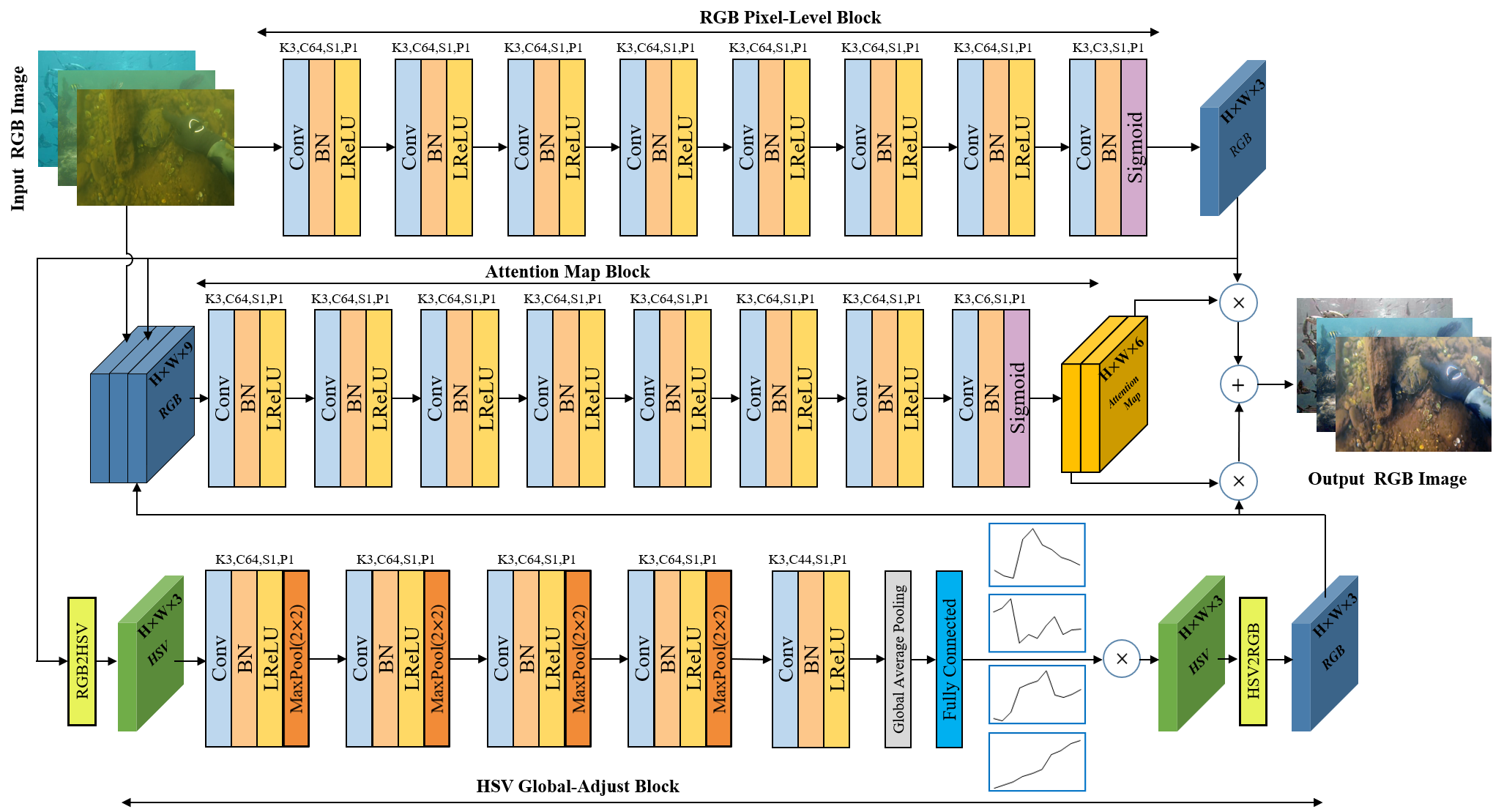}
  \caption{An overview of the proposed UIEC\^{}2-Net architecture. UIEC\^{}2-Net consists of three blocks trained end-to-end: a \textit{RGB pixel-level block} for simple and basic processing, a \textit{HSV global-adjust block} that leverages \textit{neural curve layers} for globally refine image property(saturation and luminance), and a \textit{attention map block} for getting better underwater enhanced image through attention mechanism.}
  \label{fig2}
\end{figure*}

In this paper, to make up for the above shortcomings of RGB color space based enhancement methods, we propose \textbf{UIEC\^{}2-Net}, a novel CNN-based underwater image enhancement methods, using both \textbf{RGB Color Space} and \textbf{HSV Color Space}. UIEC\^{}2-Net contains three blocks as follow: a RGB pixel-level block for simple and basic operations such as denoising and removing color cast, a HSV global-adjust block for globally adjust and fine-tune underwater image such as saturation and brightness, and an attention map block for combining the advantages of RGB pixel-level block and HSV global-adjust block output images. Similar to \cite{schwartz2018deepisp}, UIEC\^{}2-Net includes global image transformations to adjust colors, brightness, saturation, \textit{etc}. Same as \cite{li2019underwater}, the RGB pixel-level block is a plain fully CNN, which we do not use skip connection or downpooling. Although most of the processing results of the RGB block is already doing very well, there are still exists some problem such as insufficient texture details and the background color cast cannot restored well. Inspired by \cite{moran2019difar}, we applied a HSV global-adjust block to enhance the properties of underwater images through the learned global adjustment curves in HSV color space, which more expressively adjust underwater image properties and make the underwater image enhancement effect better, no matter on visual effect or scores of image quality evaluation standard. We simplified the model of \cite{moran2019difar}, but it still shows good performance in underwater image enhancement. However, because the H-channel of HSV color space is very sensitive, when HSV color space is used to process some underwater images, this may cause color distortion in some areas of the image. To solve the problem and further improve the quality of enhanced image, attention map block is used to combine the advantages of the output results of the above two blocks and avoid the related disadvantages at the same time. Therefore, attention map block has been proposed to learn the weight of each pixel for the output of each block. This is similar to \cite{hu2018squeeze}, but we are pixel-level instead of channel-level for better results. Our contributions in this paper are three-fold:

\begin{itemize}
  \item \textbf{An end-to-end convolution neural network for underwater image enhancement:} The convolution neural network architecture incorporating a RGB pixel-level block for simple and basic operations such as denoising and removing color cast, a HSV global-adjust block for adjusting global color cast and fine-tuning underwater image such as saturation and brightness, and a attention map block for getting better underwater enhanced image. UIEC\^{}2-Net parameters are much less than deep CNN models, meanwhile, our model has better generalization ability and gets better results on real-world underwater image datasets.
  \item \textbf{HSV global-adjust block:} We introduce a simple neural network curve layer, that learns a piece-wise linear scaling curve to adjust image properties in HSV color space, especially the saturation and brightness of the underwater image.
  \item \textbf{Two color space:} We apply a differentiable transformations of the image in HSV color spaces. To our best knowledge, this method is the first to use HSV color space for underwater image enhancement based on deep learning.

\end{itemize}

\section{Relate Work}

\textbf{Underwater image enhancement based on deep learning:} Exploring the underwater world has become a popular direction in recent years \cite{han2018review, cui2017extended}. Underwater image enhancement as an important pretreatment process aim to improve the visual quality of low-quality images. A variety of methods based on deep learning have been proposed and can be organized into two main categories, which are CNN-based algorithms and GAN-based algorithms. Moreover, the deep learning-based underwater image enhancement can be divided into 5 methods\cite{anwar2020diving} based on their architectural difference: encoder-decoder models \cite{sun2018deep,uplavikar2019all,fabbri2018enhancing}, which is a famous architecture in many low-level tasks to advance image quality; designed a modular or block has the same structure and are repeatedly applied in the network to improve the feature extraction capability of the network\cite{anwar2018deep,guo2019underwater}; designed multiple branch aims to learn different features through separate branches\cite{wang2017deep, li2019underwater,li2019fusion}; improving the performance of enhancement and restoration by predicted the depth map or transmission map of the underwater image\cite{hou2018joint, cao2018underwater,li2017watergan}; employed multiple generators to inference the improved image\cite{li2018emerging, lu2019multi, ye2018underwater}.

\textbf{Underwater image dataset:} Deep learning-based methods always need to be driven by large datasets. Unlike other low-level vision tasks such as image super-resolution\cite{li2019feedback}, image denoising\cite{zhang2017beyond}, image deblurring\cite{zhang2018dynamic} and image dehazing\cite{dong2020multi}, which can use plenty of synthetic degraded images and high-quality counterparts for training, it is difficult to synthesize realistic underwater images for deep learning training, due to the complex underwater image formation models are affected by many factors(\textit{e.g.} turbidity and lighting conditions). Therefore, it is necessary to synthesize underwater images, using GAN to synthesize low-quality underwater images is a common way to obtain paired image, Li \textit{et al.}\cite{li2017watergan} proposed a GAN-based method to synthesize underwater images from RGB-D in-air images and Fabbri \textit{et al.} \cite{fabbri2018enhancing} used Cyclegan\cite{zhu2017unpaired,guo2019underwater} to degrade the underwater image quality. Meanwhile, using underwater image formation model has gradually attracted attention recently. Blasinski \cite{blasinski2017underwater} provided a three-parameter underwater image formation model \cite{blasinski2016three} for underwater image simulation. Anwar \textit{et al.}\cite{anwar2018deep} incorporate a new underwater image synthesis method that simulates 10 different categories underwater images using NYU-v2 indoor dataset\cite{silberman2012indoor}, which is used in related research\cite{uplavikar2019all}. Although the synthetic image is similar to the real-world underwater image, there still exists a gap between them. Other methods include Duarte \textit{et al.}\cite{duarte2016dataset} simulated underwater image degradation using milk, chlorophyll, or green tea in a tank to get paired-dataset. More recently, Li \textit{et al}. constructed a real-world underwater image enhancement dataset, including 950 underwater images, 890 of which have the correspond reference images. These potential reference images are produced by 12 enhancement methods, and voted by 50 volunteers to select the final references.

\textbf{Underwater image enhancement using multiple color spaces:} Using multiple color spaces for image enhancement is a popular research direction in recent years. However, underwater image enhancement methods using multiple color spaces are all focus on conventional algorithms. Iqbal \textit{et al.}\cite{iqbal2007underwater,iqbal2010enhancing} used histogram stretching process on two different color spaces. They first use RGB color space to correct underwater image contrast and then use HSI color space to further improve the image quality. Hitam \textit{et al.}\cite{hitam2013mixture} used CLAHE on RGB and HSV color spaces and then combined Euclidean Norm to obtain enhanced image, the results shows that the image quality can be improved by enhancing the contrast. Ghani \textit{et al.}\cite{ghani2015enhancement} combined global and local contrast correction by using RGB and HSV two color spaces to enhance underwater image quality. In 2015, Ghani \textit{et al.}\cite{ghani2015underwater} used RGB and HSV color spaces during histogram stretching process, and in 2017, Ghani \textit{et al.}\cite{ghani2017automatic} proposed a method called Recursive Adaptive Histogram Modification (RAHM) using both RGB and HSV color space to improve the visual quality of the underwater images. In view of above introduction, using multiple color space to enhance underwater image shows better performance. Meanwhile, CNN has better enhancement effect than traditional methods. Therefore, in this paper, we are the first using two color spaces(RGB and HSV color space) in deep learning-based underwater image enhancement to get higher quality images. As far as we know, using multiple color spaces is not currently applied to deep learning-based underwater image enhancement methods.

\section{Proposed Model}

In this section, we first discuss the details of the proposed CNN-based underwater image enhancement using 2 color spaces(RGB and HSV color space). Then we introduce the loss function used in UIEC\^{}2-Net. Finally, we present the RGB color space and the HSV color space conversions to permit end-to-end learning via stochastic gradient descent(SGD) and backpropagation.

\subsection{Network Architecture}
UIEC\^{}2-Net is an end-to-end trainable neural network that consists of three blocks as shown in Figure \ref{fig2}. The \textit{RGB pixel-level block} is a CNN-based network for simple and basic processing such as denoising and removing color cast, the \textit{HSV global-adjust block} employing a novel neural curve layer that globally refine image properties, especially the saturation and luminance, and the \textit{attention map block} distribute weight to result of RGB and HSV block at pixel-level through the attention mechanism to obtain better enhanced underwater images.

\textbf{RGB pixel-level block:} The architecture of the RGB pixel-level block and parameter settings are shown at top of Figure \ref{fig2}. Because, downsampling may cause some problems such as missing image information, especially, for pixel-level computer vision tasks, it needs to be used in conjunction with upsampling, and needs various of convolution to offset the impact of downsampling and upsampling. We designed the RGB pixel-level block as a plain fully CNN without downsampling. It consists with eight $3 \times 3$ convolutional layers with stride 1, each of them using batch normalization and the first 7 layers followed by a Leaky ReLU activation, the last layer using sigmoid as the activation function, its purpose is to make the output in [0,1]. RGB pixel-level block produces a result with output shape $H \times W \times 3$, where $H, W$ is the height and width of feature maps, the number of channels is 3, and the output is passed over to the HSV global-adjust block for further processing. The performance of the RGB pixel-level block architecture we proposed in this paper has reached a high level, however, we believe that the widely used backbones for pixel-level tasks such as the U-Net architecture \cite{ronneberger2015u} and the residual network architecture \cite{he2016deep} can further improve the performance, we will try in the future work.

\begin{figure}[htbp]
  \centering
  \subfigure{
  \includegraphics[width=0.45\linewidth]{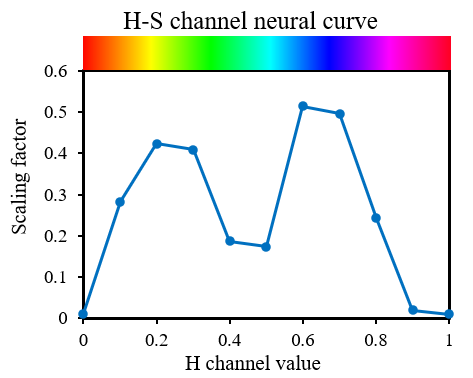}
  \includegraphics[width=0.45\linewidth]{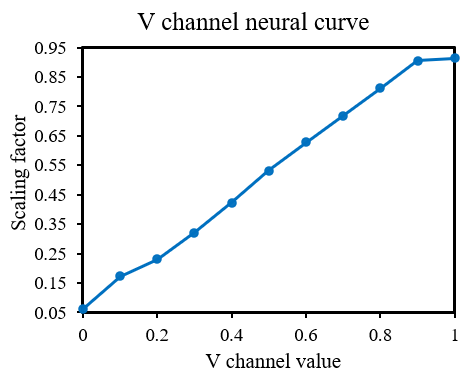}
  }
  \caption{Examples of learnt neural global adjustment curves in HSV color space.}
  \label{fig3}
\end{figure}

\textbf{HSV global-adjust block:} The output of RGB pixel-level block is transformed to HSV color space through differentiable RGB$\rightarrow$HSV and is used as input of HSV global-adjust block. Our proposed HSV global-adjust block and parameter settings are shown at the bottom of Figure \ref{fig2}, which consists of five $3 \times 3$ convolutional layers with stride 1, each of them followed by a Leaky ReLU activation. There is a $2 \times 2$ maxpooling after the first four layers and the remaining one is not. Then we place a global average pooling layer processes the shape of the feature maps to $1 \times 1 \times C$ and followed by a fully connected layer that regress the knot points of a piece-wise linear curve. The curve adjusts the predicted image(${{\hat I}_i} \in [0,1]$) by scaling pixels with the formula presented in Equation \ref{eq1}:
\begin{equation}\label{eq1}
  S(\hat I_i^{jl}) = {k_0} + \sum\limits_{m = 0}^{M - 1} {({k_{m + 1}} - {k_m})\delta (M\hat I_i^{jl} - m)}
\end{equation}
where
$$\delta \left( x \right) = \left\{ \begin{array}{l}
0{,\qquad}x < 0\\
x{,\qquad}0 \le x \le 1\\
1{,\qquad}x > 1
\end{array} \right.$$
where $M$ is the number of predicted knot points, $\hat I_i^{jl}$ is the $j$-th pixel value in the $l$-th color channel of the $i$-th image, $k_m$ is the value of the knot point $m$. The HSV global-adjust block can be seen as a simple matter of multiplication of a pixel's value: the pixel value is multiplied by the value corresponding to the piece-wise linear curve to get the globally refined images. Examples of the piece-wise linear curve that learnt by proposed model are shown in Figure \ref{fig3}.

HSV is composed of three channels(Hue, Saturation, Value), we using four curves to adjust and refine image properties. We adjust luminance by using value scaled based on value, adjust saturation by using saturation based on saturation and saturation based on hue, and refine color by using hue based on hue. The input of HSV global-adjust block is the HSV image converted from the RGB pixel-level block's output. The HSV global-adjust block produces a HSV result with output shape $H \times W \times 3$ and the HSV image is converted back to RGB space via a differentiable HSV to RGB conversion.

\textbf{Attention map block:} The raw input image is concatenated with the output of RGB pixel-level block and HSV global-adjust block as the input of this block, which is shown at the middle of Figure \ref{fig2}. The architecture of attention-map block is similar to RGB pixel-level block, which consists with eight $3 \times 3$ convolutional layers, each of them using batch normalization and the first 7 layers followed by a Leaky ReLU activation, the last layer uses Sigmoid as activation function, which can set the output between 0 and 1. The output shape of attention map block is $H \times W \times 6$, where the first three channels are weights of RGB block output and the remaining three channels belong to HSV block output. Finally, multiply the attention map with the output of RGB image converted form HSV image and output of RGB block respectively, and add the two results above to get high quality enhanced underwater image, which is the UIEC\^{}2-Net final output.

\subsection{Loss Function}
The end-to-end training of UIEC\^{}2-Net is supervised by four loss components, which consists of two color space-specific terms. We defined ${{I_i}}$ is the groundtruth image and ${{{\hat I}_i}}$ is the predicted image. Our total loss function is presented in Equation \ref{eq2}:
\begin{equation}\label{eq2}
\begin{split}
  {\cal L} = &({\lambda _{pixel}} + {\lambda _{whole}})({\omega _{\ell 1}}{{\cal L}_{\ell 1}} + {\omega _{SSIM}}{{\cal L}_{SSIM}}) + \\
  &{\omega _{hsv}}{{\cal L}_{hsv}} + {\omega _{perc}}{\cal L}_j^\phi
\end{split}
\end{equation}
where ${{\cal L}_{hsv}}$ is the HSV color space loss function, and ${{\cal L}_{\ell 1}}$, ${{\cal L}_{SSIM}}$, ${\cal L}_j^\phi$ are RGB color space loss functions. ${\lambda _{pixel}}$ represent the weight applied in RGB pixel-level block and ${\lambda _{whole}}$ is the weight applied in whole newtwork. These functions and terms are defined in more detail below.

\textbf{HSV loss(${{\cal L}_{hsv}}$):} HSV image ${{I_i}}$ is divided into three channels: hue ${H_i} \in [0,2\pi )$, saturation ${S_i} \in [0,1]$, value ${V_i} \in [0,1]$.Inspired by \cite{moran2019difar}, we compute ${{\cal L}_{hsv}}$ in the conical HSV color space:

\begin{equation}\label{eq3}
  {{\cal L}_{hsv}} = {\left\| {{{\hat S}_i}{{\hat V}_i}\cos ( {{{\hat H}_i}} ) - {S_i}{V_i}\cos \left( {{H_i}} \right)} \right\|_1}
\end{equation}
Compared to RGB, HSV has more advantages because it separates color into useful components(hue, saturation, value). It can globally adjust luminance through value-channel, improve saturation through saturation-channel and refine color through hue-channel.

\textbf{\textit{L1} loss(${{\cal L}_{\ell 1}}$):} We use $L_1$ loss on RGB pixels between the predicted and groundtruth images. This loss function is first applied in RGB pixel-level block, which computed $L_1$ distance between the predicted RGB image and groundtruth image. Then it applied to compute the distance between the UIEC\^{}2-Net output and groundtruth image. The \textit{L1} losss presented in Equation \ref{eq4}:

\begin{equation}\label{eq4}
  {{\cal L}_{\ell 1}} = {\left\| {{{\hat I}_i} - {I_i}} \right\|_1}
\end{equation}

\textbf{SSIM loss(${{\cal L}_{SSIM}}$):} We use the SSIM loss \cite{zhao2016loss} in our objective function to impose the structure and texture similarity on the predicted image. In this paper, we use gray images, which convert from RGB images, to compute SSIM score, and for each pixel $x$, the SSIM value is computed within a $11\times11$ image patch around the pixel. The formula is as follows:
\begin{equation}\label{eq5}
  \small
  SSIM(x) = \frac{{2{\mu _I}(x){\mu _{\hat I}}(x) + {c_1}}}{{\mu _I^2(x) + \mu _{_{\hat I}}^2(x) + {c_1}}} \cdot \frac{{2{\sigma _{I\hat I}}(x) + {c_2}}}{{\sigma _I^2(x) + \sigma _{_{\hat I}}^2(x) + {c_2}}}
\end{equation}
where ${{\mu _{\hat I}}(x)}$ and ${\sigma _{_{\hat I}}(x)}$ correspond to the mean and standard deviation of the RGB patch from the predicted image, similarly, ${{\mu _I}(x)}$ and ${\sigma _I(x)}$ are the same corresponds to groundtruth image. And ${{\sigma _{I\hat I}}(x)}$ is cross-covariance. We set $c_1=0.02$ and $c_2=0.03$. The SSIM loss is expressed as:

\begin{equation}\label{eq6}
  {{\cal L}_{SSIM}} = 1 - \frac{1}{N}\sum\limits_{i = 1}^N {SSIM\left( {{x_i}} \right)}
\end{equation}
The SSIM loss is similar to \textit{L1} loss, we applied it twice during training.

\textbf{Perceptual loss(${\cal L}_j^\phi$):} Inspired by \cite{johnson2016perceptual}, we defined the perceptual loss based on VGG network \cite{simonyan2014very}, which pre-trained on the ImageNet dataset\cite{deng2009imagenet}.Due to the deep layer has a better understanding of semantic information and can fully preserve the image content and overall spatial structure, and the shallow layer is more sensitive to color and texture, we select layer 4$\_$3 from VGG-19 to make it sensitive to both color and semantics. The perceptual loss expressed as the distance between the feature representations of predicted RGB image and groundtruth image:

\begin{equation}\label{eq7}
  {\cal L}_j^\phi  = \frac{1}{{{C_j}{H_j}{W_j}}}\sum\limits_{i = 1}^N {\left\| {{\phi _j}({{\hat I}_i}) - {\phi _j}({I_i})} \right\|}
\end{equation}
where ${{\phi _j}}$ is the $j$-th convolutional layer of pre-trained VGG-19; $N$ is the number of each batch in the training procedure; ${{C_j}{H_j}{W_j}}$ represents the $j$-th layer dimension of the feature maps of the VGG-19 network, which ${{C_j}, {H_j}, {W_j}}$  represent number, height and width of the feature map, respectively. The perceptual loss computed the distance between the UIEC\^{}2-Net output RGB image and groundtruth image.

\textbf{Loss term weights:} Each loss term has a weight hyperparameter: ${\omega _{\ell 1}}$,${\omega _{SSIM}}$,${\omega _{hsv}}$,${\omega _{perc}}$, meanwhile, the \textit{L1} and SSIM loss has two hyperparameters: ${\lambda _{pixel}}$,${\lambda _{whole}}$. We set ${\omega _{\ell 1}=1}$,${\omega _{SSIM}=1}$,${\omega _{hsv}}=1$,${\omega _{perc}}=0.5$, and at the first 20 epoches we set ${\lambda _{pixel}=0.5}$,${\lambda _{whole}=0.5}$,and set ${\lambda _{pixel}=0.1}$,${\lambda _{whole}=0.9}$ for the last epoches.

\subsection{Color Space Transformations}
UIEC\^{}2-Net relies on differentiable RGB$\rightarrow$HSV and HSV$\rightarrow$RGB color space conversions to permit the end-to-end learning via backpropagation. The related color space conversion methods can be found on the OpenCV website \footnote{\url{https://docs.opencv.org/3.3.0/de/d25/imgproc_color_conversions.html}}.

\textbf{RGB$\rightarrow$HSV:} The color space transformation contains minimum and maximum operations, each of them can differentiable. The RGB$\rightarrow$HSV conversion functions shown as Equations \ref{eq8}-\ref{eq10}:

\begin{equation}\label{eq8}
  V = \max (R,G,B)
\end{equation}

\begin{equation}\label{eq9}
  S = \left\{ \begin{array}{l}
\frac{{V - \min ( {R,G,B} )}}{V},\qquad if \ V \ne 0\\
0,\qquad \qquad \qquad \ \ \ otherwise
\end{array} \right.
\end{equation}

\begin{equation}\label{eq10}
  H = \left\{ \begin{array}{l}
\frac{{60( {G - B} )}}{{( {V - \min ( {R,G,B} )} )}},\qquad if\ V = R\\
\frac{{120 + 60( {B - R} )}}{{( {V - \min ( {R,G,B} )} )}},\qquad if\ V = G\\
\frac{{240 + 60( {R - G} )}}{{( {V - \min ( {R,G,B} )})}},\qquad if\ V = B
\end{array} \right.
\end{equation}
if $H < 0$ then $H = H + 360$, $0 \le V \le 1,0 \le S \le 1,0 \le H \le 360$. The operation involves conditional statements, which conditioned on the values of R, G and B. However, these can be processed under the pytorch\footnote{\url{https://pytorch.org/}} framework, and the operations such as backpropagation can be performed.

\begin{figure}[htbp]
  \centering
  \includegraphics[width=0.8\linewidth]{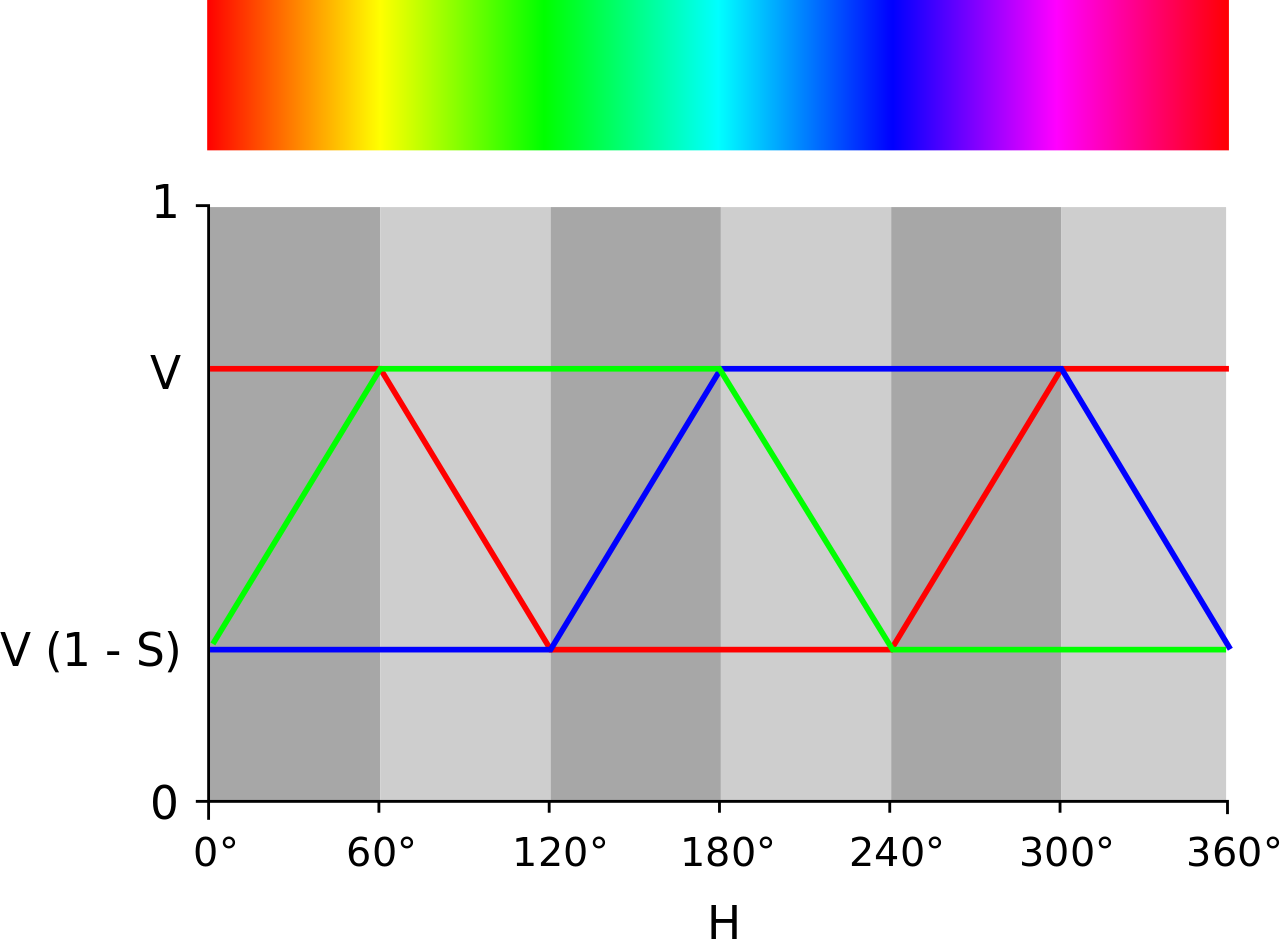}
  \caption{Piecewise linear functions used to convert HSV to RGB color space. Figure source: \url{https://commons.wikimedia.org/w/index.php?curid=1116423.}}
  \label{fig4}
\end{figure}

\textbf{HSV$\rightarrow$RGB:} The conversion rarely applied in relevant deep learning-based methods, due to the related formulas are hard to differentiable. Inspired by \cite{moran2019difar}, we replaced complex formulas with piecewise linear functions based on Figure \ref{fig4}. The corresponding piecewise linear curves are defined by linear segments and knot points, which makes them have a gradient between the points. The formulas shown as:
\begin{align}
\small
  R( {\hat I_i^j} ) =& {v_j} - ( {( {{v_s}{v_j}} )/60} )\delta ( {360{h_j} - 60} )\\  \notag
  &+( {( {{v_s}{v_j}} )/60} )\delta ( {360{h_j} - 240} )\\ \label{eq11}
  B( {\hat I_i^j} ) = & {v_j}( {1 - {s_j}} ) + ( {( {{v_s}{v_j}})/60} )\delta ( {360{h_j} - 120} ) \\ \notag
  &-( {( {{v_s}{v_j}} )/60} )\delta ( {360{h_j} - 300} ) \\ \label{eq12}
  B( {\hat I_i^j} ) = & {v_j}( {1 - {s_j}} ) + ( {( {{v_s}{v_j}} )/60} )\delta ( {360{h_j} - 120} ) \\ \notag
  &-( {( {{v_s}{v_j}} )/60} )\delta ( {360{h_j} - 300} ) \label{eq13}
\end{align}
where we define $\delta(x)$ as in Equation \ref{eq14}
\begin{equation}\label{eq14}
  \delta ( x ) = \left\{ \begin{array}{l}
0{,\qquad \ }x < 0\\
x{,\qquad \ }0 \le x \le 40\\
60{,\qquad}x > 60
\end{array} \right.
\end{equation}
where ${h_j},{s_j},{v_j}$ are the hue, saturation and value of pixel $j$ of image $I_i$ ,respectively. and both of them belongs $[0,1]$.

\section{Experimental Results}

To evaluate our method, we perform qualitative (subjective comparisons) and quantitative (subjective comparisons) comparisons with traditional methods and the recent state-of-the-art deep learning-based underwater image enhancement methods on both synthetic and real-world underwater images. These methods include Histogram Equalization(HE), White Balance(WB), UDCP\cite{drews2016underwater}, ULAP\cite{song2018rapid}, UWGAN\cite{wang2019uwgan}, UGAN\cite{fabbri2018enhancing} ,UWCNN\cite{anwar2018deep}, DUIENet\cite{li2019underwater}. For deep learning-based methods, we run the source codes with the pre-trained model parameters shared by provided by corresponding authors to produce the best results for an objective evaluation. In this section, we will first supplement the training details, and then analyzed the experimental results of synthetic and real-world underwater images.

\subsection{Implementation Details}

For training, the input of our network are both synthetic and real-world underwater images. A random set of 800 pairs of real-world images extracted from the UIEBD dataset \cite{li2019underwater} and 1200 pairs of synthetic images with ten types are generated from the NYU-v2 RGB-D dataset based on \cite{anwar2018deep}, both of them (2000 images in total) are used to train our network. We resize the input images to size $350 \times 350$ and randomly crop them to size $320 \times 320$. For testing, 90 real-world and 900 synthetic images are treated as the testing set. Compared to training, we do not resize or randomly crop the input image. We trained our model using ADAM and set the learning rate to 0.0001. The batch size is set to 8 and the epoch is set to 50. We use Pytorch as the deep learning framework on an Inter(R) i7-8700k CPU, 32GB RAM, and a Nvidia GTX 1080Ti GPU.

\subsection{Experiment on Synthetic Datasets}
\label{sec1}
\begin{figure*}[htbp]
  \centering
  \setlength{\belowcaptionskip}{0pt}
  \setlength{\abovecaptionskip}{0pt}
  \setcaptionwidth{6.5in}
  \includegraphics[width=0.9\linewidth]{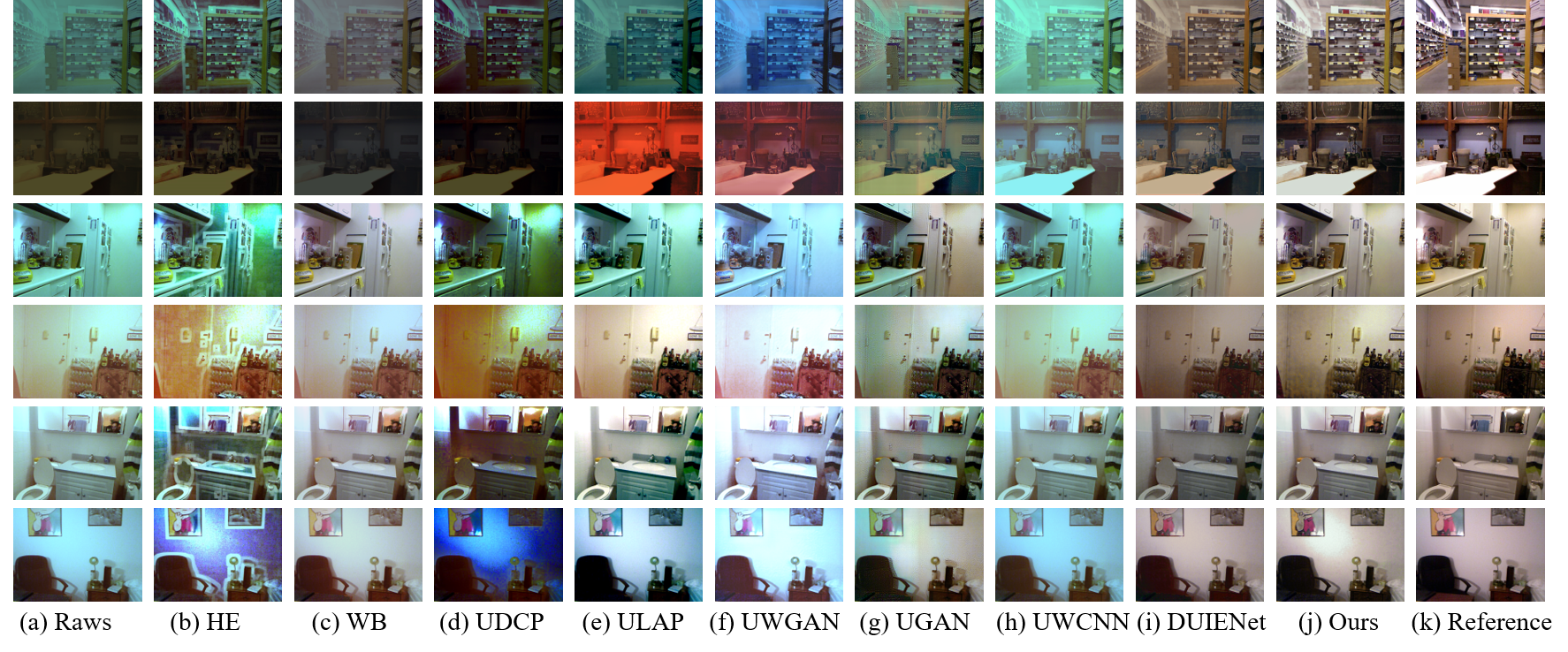}
  \caption{Subjective comparisons on synthetic underwater images. From left to right are raw underwater images, the results of Histogram Equalization(HE), White Balance(WB), UDCP\cite{drews2016underwater}, ULAP\cite{song2018rapid}, UWGAN\cite{wang2019uwgan}, UGAN\cite{fabbri2018enhancing} ,UWCNN\cite{anwar2018deep}, DUIENet\cite{li2019underwater}, the proposed UIEC\^{}2-Net and reference images.}
  \label{fig5}
\end{figure*}

It is very common to train the network through synthetic underwater images because there is no GT for underwater images, thus, We first evaluate the capacity of our method on synthetic testing set.
In Figure \ref{fig5}, we present the results of underwater image enhancement on the synthetic underwater images from our testing set.

\begin{figure*}[htbp]
  \centering
  \setlength{\belowcaptionskip}{0pt}
  \setlength{\abovecaptionskip}{0pt}
  \setcaptionwidth{6.5in}
  \includegraphics[width=0.9\linewidth]{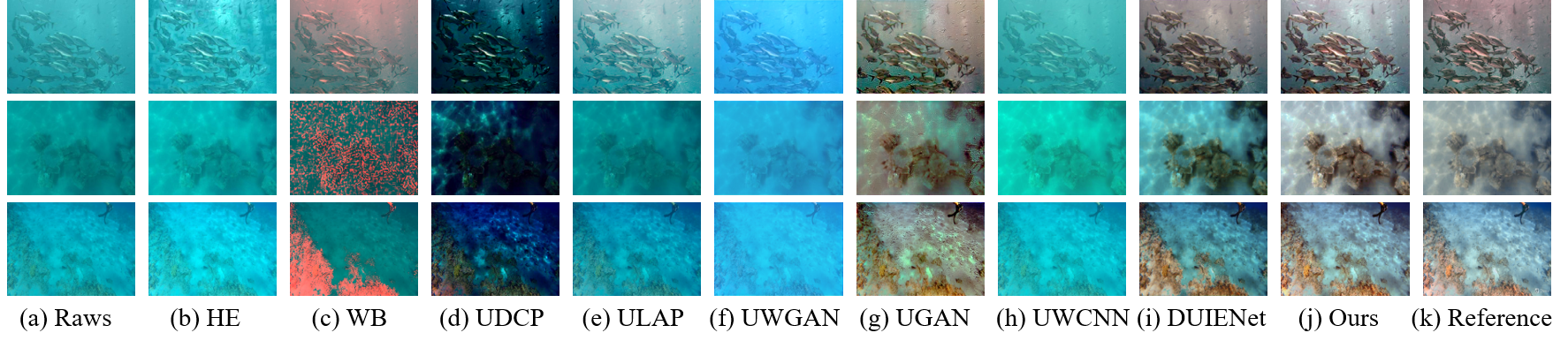}
  \caption{Subjective comparisons on bluish underwater images. From left to right are raw underwater images, the results of Histogram Equalization(HE), White Balance(WB), UDCP\cite{drews2016underwater}, ULAP\cite{song2018rapid}, UWGAN\cite{wang2019uwgan}, UGAN\cite{fabbri2018enhancing} ,UWCNN\cite{anwar2018deep}, DUIENet\cite{li2019underwater}, the proposed UIEC\^{}2-Net and reference images.}
  \label{fig6}
\end{figure*}
\begin{figure*}[htbp]
  \centering
  \setlength{\belowcaptionskip}{0pt}
  \setlength{\abovecaptionskip}{0pt}
  \setcaptionwidth{6.5in}
  \includegraphics[width=0.9\linewidth]{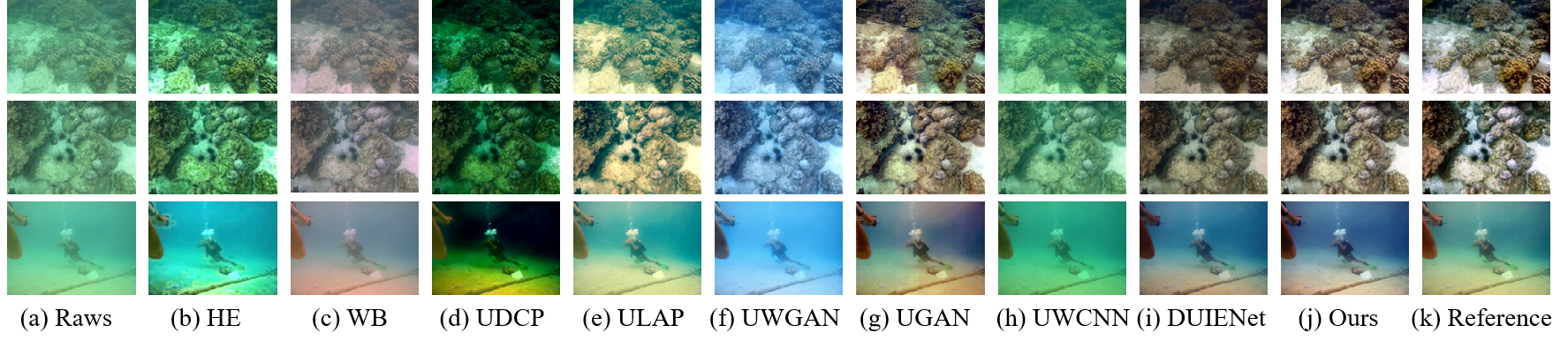}
  \caption{Subjective comparisons on greenish underwater images. From left to right are raw underwater images, the results of Histogram Equalization(HE), White Balance(WB), UDCP\cite{drews2016underwater}, ULAP\cite{song2018rapid}, UWGAN\cite{wang2019uwgan}, UGAN\cite{fabbri2018enhancing} ,UWCNN\cite{anwar2018deep}, DUIENet\cite{li2019underwater}, the proposed UIEC\^{}2-Net and reference images.}
  \label{fig7}
\end{figure*}
\begin{figure*}[htbp]
  \centering
  \setlength{\belowcaptionskip}{0pt}
  \setlength{\abovecaptionskip}{0pt}
  \setcaptionwidth{6.5in}
  \includegraphics[width=0.9\linewidth]{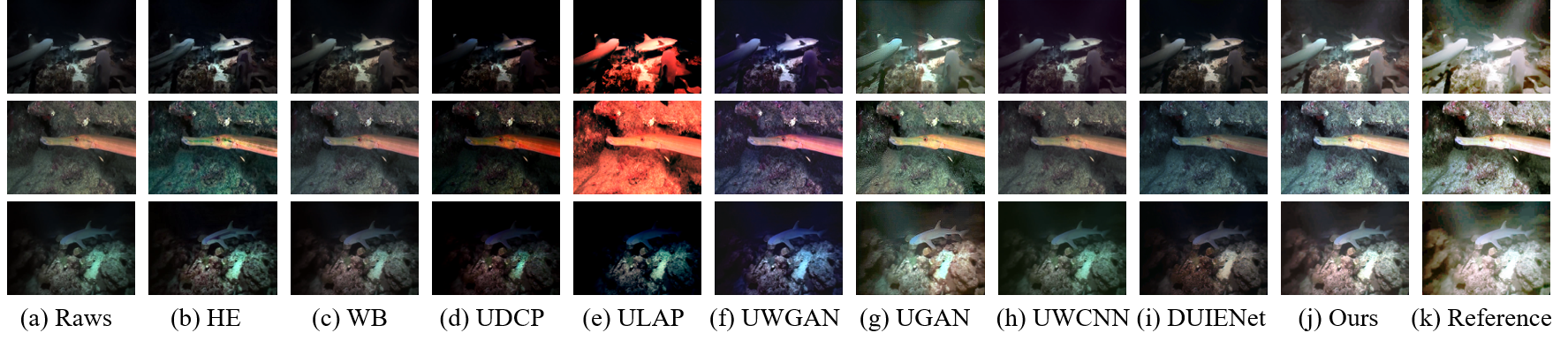}
  \caption{Subjective comparisons on  low-illuminated underwater images. From left to right are raw underwater images, the results of Histogram Equalization(HE), White Balance(WB), UDCP\cite{drews2016underwater}, ULAP\cite{song2018rapid}, UWGAN\cite{wang2019uwgan}, UGAN\cite{fabbri2018enhancing} ,UWCNN\cite{anwar2018deep}, DUIENet\cite{li2019underwater}, the proposed UIEC\^{}2-Net and reference images.}
  \label{fig8}
\end{figure*}
\begin{figure*}[htbp]
  \centering
  \setlength{\belowcaptionskip}{0pt}
  \setlength{\abovecaptionskip}{0pt}
  \setcaptionwidth{6.5in}
  \includegraphics[width=0.9\linewidth]{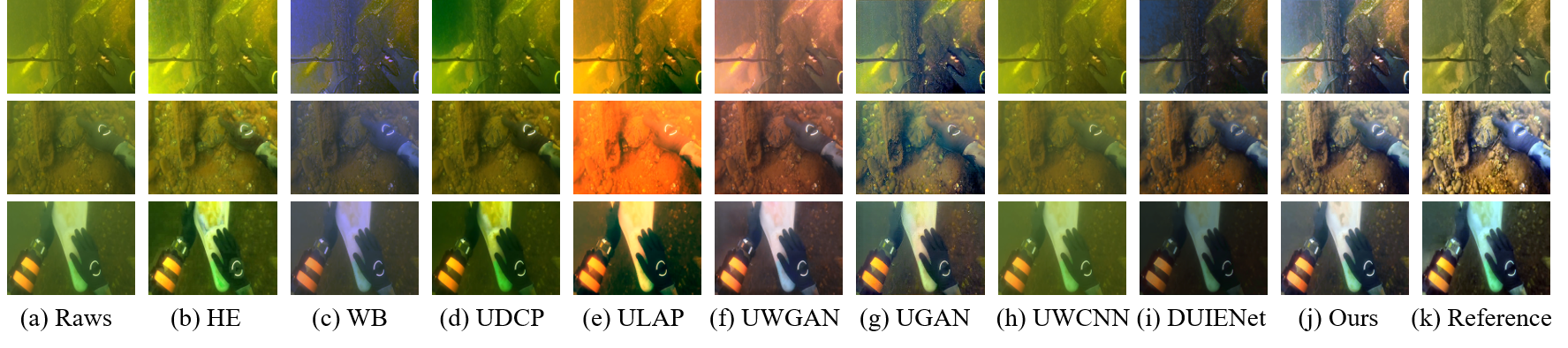}
  \caption{Subjective comparisons on yellowish underwater images. From left to right are raw underwater images, the results of Histogram Equalization(HE), White Balance(WB), UDCP\cite{drews2016underwater}, ULAP\cite{song2018rapid}, UWGAN\cite{wang2019uwgan}, UGAN\cite{fabbri2018enhancing} ,UWCNN\cite{anwar2018deep}, DUIENet\cite{li2019underwater}, the proposed UIEC\^{}2-Net and reference images.}
  \label{fig9}
\end{figure*}
\begin{figure*}[htbp]
  \centering
  \setlength{\belowcaptionskip}{0pt}
  \setlength{\abovecaptionskip}{0pt}
  \setcaptionwidth{6.5in}
  \includegraphics[width=0.9\linewidth]{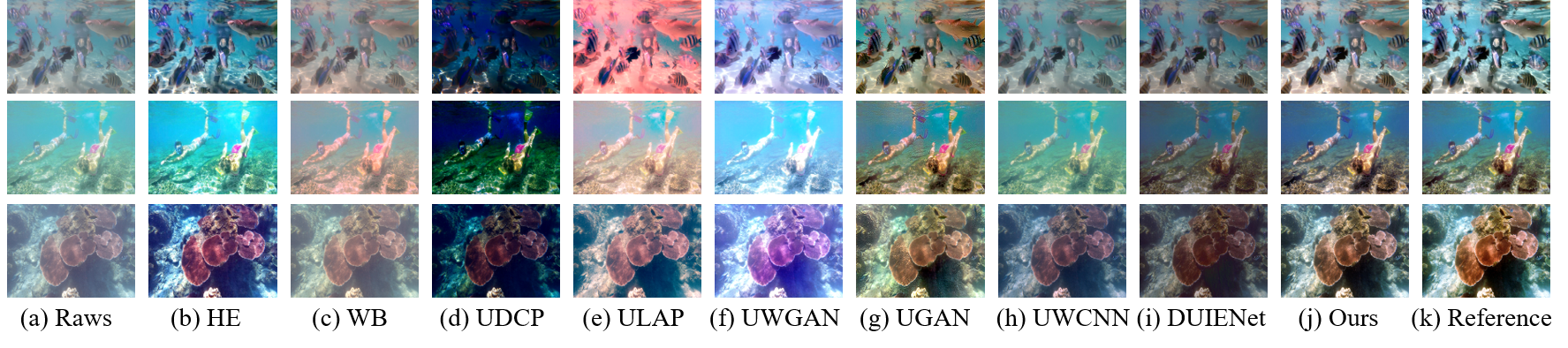}
  \caption{Subjective comparisons on shallow water images. From left to right are raw underwater images, the results of Histogram Equalization(HE), White Balance(WB), UDCP\cite{drews2016underwater}, ULAP\cite{song2018rapid}, UWGAN\cite{wang2019uwgan}, UGAN\cite{fabbri2018enhancing} ,UWCNN\cite{anwar2018deep}, DUIENet\cite{li2019underwater}, the proposed UIEC\^{}2-Net and reference images.}
  \label{fig10}
\end{figure*}
\captionsetup[table]{name={Table},labelfont={sc},labelsep=newline}
\begin{table}[htbp]
  \setlength{\belowcaptionskip}{0pt}
  \setlength{\abovecaptionskip}{-5pt}
  \centering
  \caption{}
  \begin{center}
      \textsc{Full-reference Image Quality Assessment in Terms of MSE, \newline{PSNR and SSIM on Synthetic Images.}}
  \end{center}
  \begin{tabular}{c|c|c|c}
    \hline
        \textbf{Method} & \textbf{MSE} & \textbf{PSNR(dB)} & \textbf{SSIM} \\
    \hline
    raw	                              &  4142.5386	&  14.4978  & 	0.6948	\\	
    HE	                              &  3869.0180	&  14.2563  & 	0.6779 \\
    WB	                              &  4246.7734	&  14.5174  & 	0.6862 \\
    UDCP\cite{drews2016underwater}    &  4809.4566	&  13.6778  & 	0.6373 \\    ULAP\cite{song2018rapid}	      &  3725.4596	&  14.3739  & 	0.6989 \\
    UWGAN\cite{wang2019uwgan}	      &  4028.8004	&  13.3007  & 	0.7083 \\
    UGAN\cite{fabbri2018enhancing}    &  \textcolor{blue}{2266.4678}	&  \textcolor{blue}{16.2636}  & 	0.6625 \\
    UWCNN\cite{anwar2018deep}         &  3770.7036	&  14.8638  & 	0.7419 \\
    DUIENet\cite{li2019underwater}    &  2323.3417	&  16.1073  & 	\textcolor{blue}{0.7734} \\
    UIEC\^{}2-Net	                  &  \textcolor{red}{1126.3743}  &  \textcolor{red}{20.5442}  & 	\textcolor{red}{0.8486} \\
    \hline
  \end{tabular}
\label{table1}
\end{table}
As shown in Figure \ref{fig5}(a),we simply give examples of synthetic underwater images, generated from RGB-D in-air images. Figure \ref{fig5}(b)-(k) shows the reference image of Histogram Equalization(HE), White Balance(WB), UDCP\cite{drews2016underwater}, ULAP\cite{song2018rapid}, UWGAN\cite{wang2019uwgan}, UGAN\cite{fabbri2018enhancing} ,UWCNN\cite{anwar2018deep}, DUIENet\cite{li2019underwater}, the proposed UIEC\^{}2-Net and the corresponding reference image respectively. It is noted that UWCNN\cite{anwar2018deep} has 10 different pre-train models, which are trained separately by 10 types. We only choose type-1 model as a comparative experiment, so when the underwater images category is rich, its generalization performance is not good. Similarly, for other methods, they have better processing results in one or two categories. However, in most cases, they are not good at processing color cast and image details. The methods we proposed has good generalization performance and can restore color cast better, which Figure \ref{fig5} can demonstrating its effectiveness and robustness.

We choose Mean Square Error (MSE), Peak Signal to Noise Ratio (PSNR) and Structural SIMilarity index (SSIM) as full-reference image quality assessment to assess the processed synthetic underwater images for all of the compared methods. For MSE and PSNR metrics, the lower MSE (higher PSNR) denotes the result is more close to the GT in terms of image content. For SSIM, the higher SSIM scores mean the result is more similar to the GT in terms of image structure and texture. Table \ref{table1} reports the quantitative results of different methods on the synthetic testing images. We highlight the top 1 performance in \textcolor{red}{red}, whereas the second best is in \textcolor{blue}{blue}. Regarding the SSIM response, UIEC\^{}2-Net is higher than the second-best performer. Similarly, our PSNR is obviously higher (less erroneous as indicated by the MSE scores) than the compared methods. This can demonstrate that our proposed method achieves the best performance in terms of full-reference image quality assessment on synthetic underwater images.

\subsection{Experiment on Real Datasets}

From test dataset, we extracted five categories underwater images (greenish underwater images, bluish underwater images, yellowish underwater images, shallow water images, and underwater images image with limited illumination) to compare the generalization capability of underwater enhancement methods. Note that our test dataset and corresponding reference images are provided by UIEBD\cite{li2019underwater}. Due to the nature of light propagation, the red light first disappears in water, followed by green light and then the blue light. Most underwater images are bluish and greenish, such as the raw underwater images in (a) of Figure \ref{fig6} and \ref{fig7}. Underwater images also include low-brightness situations(shown in Figure \ref{fig8}), yellowish styles (shown in Figure \ref{fig9}) and shallow water images(shown in Figure \ref{fig10}).
As shown in Figure \ref{fig6}-\ref{fig10}, color deviation seriously affects the visual quality of underwater images. Traditional methods usually consider only one type of underwater images. HE effectively improves the contrast but cannot remove the color cast well. WB can improve the greenish underwater images by supplementing red light, but the results are not good in other cases. ULAP\cite{song2018rapid} can better enhance the greenish underwater images due to a good estimate of the underwater optical attenuation but the bluish underwater image cannot be improved well. UDCP\cite{drews2016underwater} aggravate the effect of the color cast. Deep learning-based methods have poor generalization and low sensitivity to brightness and saturation of underwater images, and tend to introduce artifacts, over-enhancement and color casts. By contrast, (k) of Figure \ref{fig6}-\ref{fig10} shows that our proposed UIEC\^{}2-Net effectively removes the haze and color casts (especially the background color cast) of the underwater images, adjusts underwater image properties (\textit{e.g.} brightness and saturation) and has good generalization capability in dealing with underwater images. In addition, our results even achieve better visual quality than the corresponding reference images(\textit{e.g.} less noise and better details).
\begin{table}
  \centering
  \setlength{\belowcaptionskip}{0pt}
  \setlength{\abovecaptionskip}{-5pt}
  \caption{}
  \begin{center}
      \textsc{Full-reference Image Quality Assessment in Terms of MSE, \newline{PSNR and SSIM on Real-World Images.}}
  \end{center}

  \begin{tabular}{c|c|c|c}
    \hline
    \textbf{Method} & \textbf{MSE} & \textbf{PSNR(dB)} & \textbf{SSIM} \\
    \hline
    raw	                            & 1322.1355	    & 18.2701	& 0.8151 \\
    HE	                            & 1078.9476	    & 19.5854	& \textcolor{blue}{0.8509} \\
    WB	                            & 1455.7350	    & 17.9261	& 0.8041 \\
    UDCP\cite{drews2016underwater}  & 5829.6013	    & 11.1646   & 0.5405 \\
    ULAP\cite{song2018rapid}	    & 1517.6039	    & 18.6789	& 0.8194 \\
    UWGAN\cite{wang2019uwgan}	    & 1256.0906	    & 18.6209	& 0.8454 \\
    UGAN\cite{fabbri2018enhancing} 	& \textcolor{blue}{558.2965}	& \textcolor{blue}{21.3031}	& 0.7691 \\
    UWCNN\cite{anwar2018deep}       & 1342.7639	& 18.2851	& 0.8150 \\
    DUIENet\cite{li2019underwater}	& 2023.3417	&16.2906	& 0.7884 \\
    UIEC\^{}2-Net	                & \textcolor{red}{365.5963}	& \textcolor{red}{24.5663}	& \textcolor{red}{0.9346} \\
    \hline
  \end{tabular}
\label{table2}
\end{table}
\begin{table}
  \centering
  \setlength{\belowcaptionskip}{0pt}
  \setlength{\abovecaptionskip}{-5pt}
  \caption{}
  \begin{center}
  \textsc{No-reference Image Quality Evaluation in Terms of UCIQE,\newline{UICM, UISM, UIConM and UIQM on Real-World Images.}}
  \end{center}
  \begin{tabular}{c|c|c|c|c|c}
    \hline
    \textbf{Method} & \textbf{UCIQE} & \textbf{UICM} & \textbf{UISM} & \textbf{UIConM} & \textbf{UIQM} \\
    \hline
    raw     	                    &0.5044	&2.5656	&2.7255	&0.5492	&2.8407 \\
    HE	                            &0.5828	&4.3170	&3.6235	&0.7017	&3.7006 \\
    WB	                            &0.5429	&\textcolor{red}{5.1675}	&2.6102	&0.5472	&2.8731 \\
    UDCP\cite{drews2016underwater}	&0.5747	&4.0233	&2.7689	&0.6169	&3.1367 \\
    ULAP\cite{song2018rapid}	    &0.6014	&4.9015	&2.8597	&0.5145	&2.8224 \\
    UWGAN\cite{wang2019uwgan}	    &0.5352	&3.0011	&3.1388	&0.6169	&3.2174 \\
    UGAN\cite{fabbri2018enhancing}	&\textcolor{blue}{0.6162}	&4.2476	&\textcolor{blue}{4.2527}	&\textcolor{red}{0.7384}	&\textcolor{blue}{4.0157} \\
    UWCNN\cite{anwar2018deep}	    &0.5044	&2.55656	&2.7255  &0.5492	&2.8407 \\
    DUIENet\cite{li2019underwater}	&0.6051	&4.0727	&3.8559	&0.6940	&3.7988 \\
    UIEC\^{}2-Net	                &\textcolor{red}{0.6193}	&\textcolor{blue}{4.9046}	&\textcolor{red}{4.7558}	&\textcolor{blue}{0.7094}	&\textcolor{red}{4.0790} \\
    \hline
    Reference	                    &\textbf{0.6451}	&\textbf{5.2137}	&\textbf{3.9190}	&\textbf{0.7116} &\textbf{3.8483} \\
    \hline
  \end{tabular}
\label{table3}
\end{table}

Similarly to Section \ref{sec1}, we choose MSE, PSNR and SSIM to assess the recovered results on real-world underwater images. We calculate the results of each method and the corresponding reference image, the quantitative results of different methods on real-world underwater images are shown in Table \ref{table2}. Our method achieves the best performance in terms of full-reference image quality assessment, which can prove our proposed method is good at handling details.

Meanwhile, we choose underwater color image quality evaluation (UCIQE)\cite{yang2015underwater} and underwater image quality measure (UIQM)\cite{panetta2015human} as No-reference Image Quality Evaluation. UCIQE evaluate underwater image quality by color density, saturation and contrast. UIQM is a comprehensive underwater image evaluation index, which is the weighted sum of underwater image colorfulness measure(UICM), underwater image sharpness measure(UISM) and underwater image conreast measure(UIConM):$UIQM = {c_1} \times UICM + {c_2} \times UISM + {c_3} \times UIQM$. We set $c_1=0.0282$,$c_2=0.2953$,$c_3=3.5753$ according to \cite{panetta2015human}. As shown in Table \ref{table3}, compared with other methods, our proposed method achieve the best score in UCIQE, UISM and UIQM, and even get higher score than reference image in UIQM and UISM. This also demonstrates that our algorithm has more advantages in processing details.

\begin{figure*}[htbp]
  \centering
  \setlength{\belowcaptionskip}{0pt}
  \setlength{\abovecaptionskip}{5pt}
  \setcaptionwidth{6.5in}
  \includegraphics[width=0.9\linewidth]{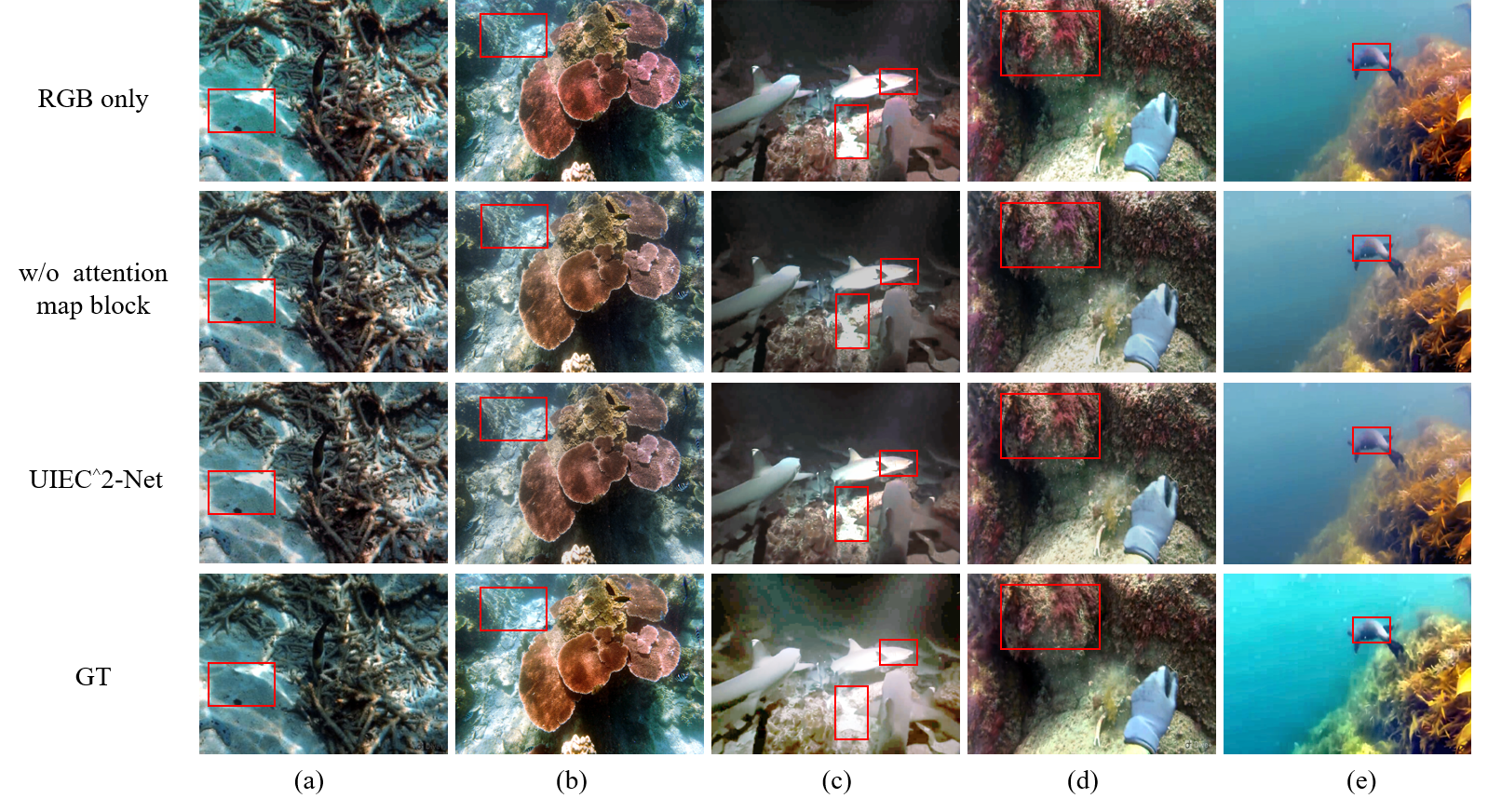}
  \caption{The enhanced result of only using RGB pixel-level block, without(w/o) using HSV global-adjust block and UIEC\^{}2-Net. Top row: the result of only use RGB pixel-level block; Middle row: the result of UIEC\^{}2-Net not using HSV global-adjust block; Bottom row: the result of UIEC\^{}2-Net.}
  \label{fig11}
\end{figure*}
\begin{table*}[htbp]
  \centering
  \setlength{\belowcaptionskip}{0pt}
  \setlength{\abovecaptionskip}{-5pt}
  \caption{}
  \begin{center}
  \textsc{Image Quality Assessment of with/without HSV Rtouching Block and Attention Map Block.}
  \end{center}
  \begin{tabular}{c|c|c|c|c|c}
    \hline
    \textbf{Method} & \textbf{MSE} & \textbf{PSNR(dB)} & \textbf{SSIM} & \textbf{UCIQE} & \textbf{UIQM}\\
    \hline
    only RGB pixel-level block &397.3553	&23.8228	&0.92849	&\textcolor{red}{0.6351}	&4.0732 \\
    w/o attention map block	   &393.9875	&24.1794	&0.92651	&0.60183	&\textcolor{red}{4.0979} \\
    UIEC\^{}2-Net              &\textcolor{red}{365.5963}	&\textcolor{red}{24.5663}	&\textcolor{red}{0.9346}	&0.6193	    &4.0790 \\
    \hline
  \end{tabular}
\label{table4}
\end{table*}
\subsection{Ablation Study}
To demonstrate the effect of HSV global-adjust block and attention map block in our network, we compared the proposed UIEC\^{}2-Net with only using RGB pixel-level block and without(w/o) attention map block as an ablation study. As show in Table \ref{table4}, although the use of HSV block and attention map block decreased performance of UCIQE and UIQM, such a sacrifice is necessary to improved whole network preference. In addition, it can achieve good results in subjective perception of underwater image.

Such an example is presented in Figure \ref{fig11}, (a),(b) demonstrate that after adding HSV global-adjust block, the background color cast of underwater image is processed better and the saturation has been improved to make the processed images more realistic. Figure \ref{fig11}(c) shows that although adopting global-adjust block can effectively avoid the luminance problem such as overexposure, HSV block may also over-processing the image to reduce the contrast and saturation of the entire image. Meanwhile, due to the H-channel of HSV color space is very sensitive, HSV block may causes color distortion in some reddish areas when processing the underwater images. These shows in the second row of Figure \ref{fig11}(d). Attention map block can solve the problem by extracting the feature information of the Raw images, results of RGB and HSV, then distributing weight of each pixel to combine the advantages of results from RGB and HSV block. As shown in (c),(d), some problems with HSV blocks can be avoided when UIEC\^{}2-Net using attention map block can. In addition, attention map block can also avoid the appearance of noise blocks, which shown in (e)

\section{Conclusion}

In this paper, we proposed a novel CNN-based underwater image enhancement method, using both RGB Color Space and HSV Color Space. It is the first attempt that using HSV color spaces in deep learning-based underwater enhancement research. We first use pixel-level block based on RGB color space for simple and basic enhancement operations such as removing color cast and denosing, then we use a global-adjust block based on HSV color space for globally refining underwater image properties such as luminance and saturation. Our method has a good effect on removing color cast, especially for the restoration and enhancement of background color, and also an greatly retain the detailed information of the underwater image. Furthermore, our method can be use as a guide for subsequent research of underwater image color correction. Experiments on synthetic and real-world underwater images including qualitative and quantitative assessment demonstrated the effectiveness of our method.

\bibliographystyle{IEEEtran}
\bibliography{mybibfile}

\end{document}